# Contrastive Learning-based Multi-Modal Architecture for Emoticon Prediction by Employing Image-Text Pairs

Ananya Pandey[1], Dinesh Kumar Vishwakarma[1*]

Biometric Research Laboratory, Department of Information Technology, Delhi Technological University, Bawana Road, Delhi-110042, India
ananyapandey_2k21phdit08@dtu.ac.in[1] , dinesh@dtu.ac.in[1*]

**Abstract-** The emoticons are symbolic representations that generally accompany the textual content to visually enhance or summarize the true intention of a written message. Although widely utilized in the realm of social media, the core semantics of these emoticons have not been extensively explored based on multiple modalities. Incorporating textual and visual information within a single message develops an advanced way of conveying information. Hence, this research aims to analyze the relationship among sentences, visuals, and emoticons. For an orderly exposition, this paper initially provides a detailed examination of the various techniques for extracting multimodal features, emphasizing the pros and cons of each method. Through conducting a comprehensive examination of several multimodal algorithms, with specific emphasis on the fusion approaches, we have proposed a novel contrastive learning-based multimodal architecture. The proposed model employs the joint training of dual-branch encoder along with the contrastive learning to accurately map text and images into a common latent space. Our key finding is that by integrating the principle of contrastive learning with that of the other two branches yields superior results. The experimental results demonstrate that our suggested methodology surpasses existing multimodal approaches in terms of accuracy and robustness. The proposed model attained an accuracy of **91%** and an MCC-score of **90%** while assessing emoticons using the Multimodal-Twitter Emoticon dataset acquired from Twitter. We provide evidence that deep features acquired by contrastive learning are more efficient, suggesting that the proposed fusion technique also possesses strong generalisation capabilities for recognising emoticons across several modes.

## 1 Introduction

In the digital age of the social media platforms and internet, an exciting new way of human interaction has emerged. It involves the combination of concise, readable text messages and imagery ideograms known as emoticons. Emoticons are tiny symbols that represent individuals, settings, and objects. These symbolic expressions have gained widespread acceptance as a standard for communication on the web [1]. It is commonly used not just on Twitter but also on other well-known platforms like YouTube, WhatsApp, Telegram, Facebook, Instagram, and LinkedIn. According to Google Trends, the popularity of emoticons has been on the rise over the last decade, as seen in **Figure 1**. Emoticon prediction based solely on text has garnered attention and has been studied extensively from the perspective of Natural Language Processing. Aoki et al. [2]; Barbieri et al. [3]; Barbieri et al. [4]; Eisner et al. [5]; Ljubesic et al. [6]; and Boutet et al. [7] are few prominent exceptions encompass research dedicated to the semantics and usage of emoticons.

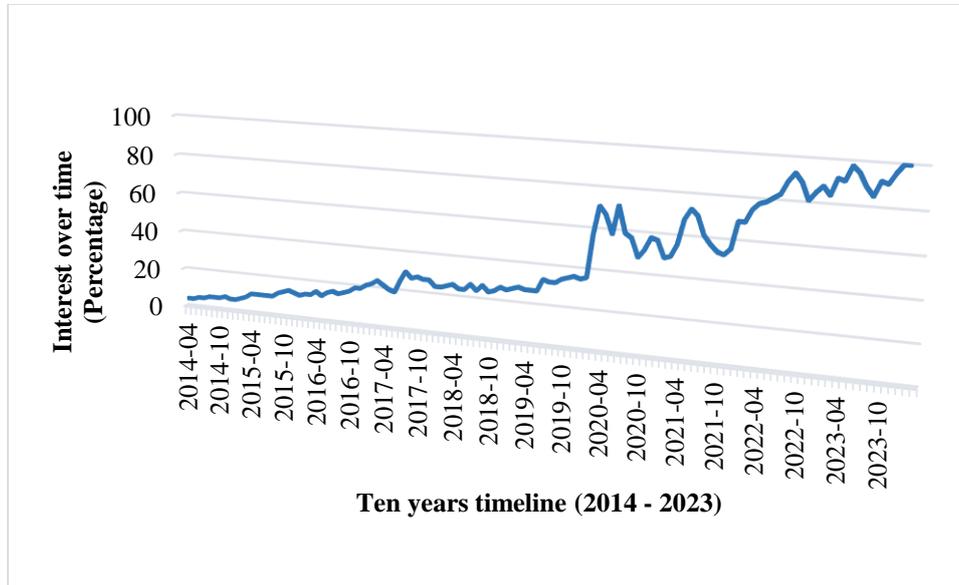

**Figure 1** Visual representation of the rise in emoticon usage over the past decade based on Google Trends data. Link

An immense amount of research has been conducted in the domain of emoticon prediction based on textual content. However, there is a dire need for further research on predicting emoticons based on multiple modalities. Hence, this research article demonstrates the importance of integrating visual information with texts in the realm of multimodal communication. Specifically, we highlight how combining texts and images in online communities can lead to more precise emoticon prediction models. We examine the utilization of emoticons within the popular social media platform Twitter. We propose a multimodal strategy to forecast the emoticons associated with a Twitter post, considering both its textual content and accompanying image. We rely on visual modality to enhance the process of selecting the most suitable emoticons for a post. Our research demonstrates that incorporating both text and images in posts enhances the precision of emoticon prediction in comparison to relying solely on textual information. It may be inferred that textual and visual content incorporate distinct yet complementary aspects of using emoticons.

In a nutshell, an efficient method for determining the correct emoticon for different types of content can benefit a wide range of applications, such as recommending emoticons for text messages, sentiment analysis [8], hate speech detection [9], humour identification, sarcasm recognition [10]-[11], generating emoticon-rich posts, etc. Considering the fact that emoticons have the potential to deceive humans, automated emoticon forecasting software might pave the way for better language understanding. Therefore, by understanding the semantics of emoticons, we may enhance highly subjective tasks like emotion, sentiment and sarcasm recognition.

| Number | I | II | III | IV |
|---|---|---|---|---|
| Textual modality | RT @lovsickgirls : lisa's reaction after she was told about her solo achievements | She is sooooo beautiful #MehndiHaiRachneWaali | RT @offl_Lawrence : A small request for my physically abled dancer boys | RT @AldrineEsther : My mother is no morehow am I going to survive oh God |
| Visual modality | 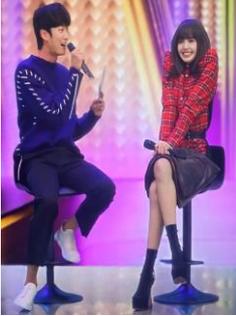 | 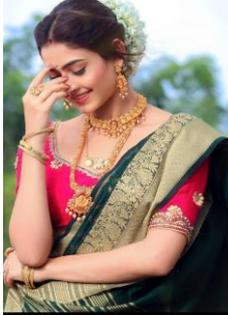 | 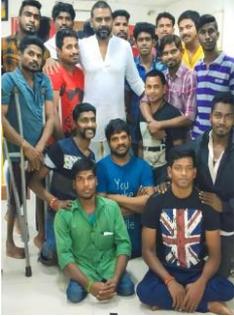 | 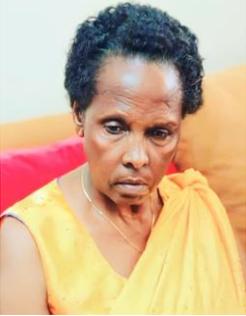 |
| Predicted Emoticon | 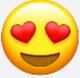 | 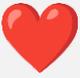 | 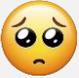 | 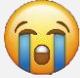 |

Numerous cutting-edge research studies have been performed, such as Barbieri et.al [12] presented the concept of emoticon prediction on Twitter, demonstrating the superiority of LSTM-based models over human performance in this task. Barbieri et.al [12] achieved promising results, but their study was limited to text-based data. Traditional research mostly focused on text-based modality. As a consequence, there is an urgent need to develop an automated approach for emoticon prediction across several modalities. As a result, this study differs from conventional emoticon forecasting research by incorporating multiple data sources instead of just text. This type of problem presents a key challenge in leveraging information from each modality for more precise analysis. Additionally, as online community statistics often include combinations of images and text, we decided to evaluate the robustness of our proposed strategy on one of the standard datasets provided by Ebrahimian et al. [13] referred as Multimodal-TwitterEmoticon, which includes image-text pairs. **Figure 2** displays few samples of the dataset to represent multimodal Twitter posts for emoticon prediction. Our research study has the following key contributions:

- We have introduced an innovative dual-branch Contrastive Learning-Based Multimodal Architecture to deal with the challenge of multimodal emoticon prediction. The proposed multimodal architecture comprises of three primary components: Transformer-based visual encoder, Transformer-based textual encoder and an additional component involves the use of contrastive learning to uncover the hidden relationships within the text and images. It has been demonstrated that by integrating the principle of contrastive learning with that of the other two branches yields superior results.
- We conduct thorough analyses and experimentation with Multimodal-TwitterEmoticon standard dataset to illustrate our model's ability to accurately and consistently model the multimodal representations of images and descriptive text. Moreover, we highlight how our model delivers ground-breaking outcomes in the realm of multimodal emoticon prediction.

- Perform an extensive study of recent literature and classify research papers related to emoticon prediction in social media analysis for both unimodal and multimodal approaches.

The article starts by outlining the motivation and summarizing the essential elements and impacts of emoticon forecasting in the introductory **Section 1**. **Section 2** provides a summary of the most recent research works on emoticon prediction using both single and multimodality approaches. The proposed technique has been succinctly explained in **Section 3**. **Section 4** presents the performance metrics used to assess the strength of our suggested technique and the results obtained from the experiment. The study concludes in **Section 5** with suggestions for future research topics.

## 2 Related Work

This section covers all the latest research studies on emoticon prediction, with an emphasis on both single and multimodal approaches. To make things simple and easy for readers to follow, all of the research outcomes are provided in a tabular manner depending on several factors in **Table 1**.

**Table 1** Summarization of latest research studies for emoticon prediction based on both single and multimodal approaches.

| Year & Ref. | Methods used for feature extraction | Fusion approach | Dataset utilized | Modality | Experimental results | Limitation |
|---|---|---|---|---|---|---|
| [14], 2018 | Trained 2 distinct classification algorithms on 400,000 tweets namely Stochastic Gradient Descent (SGD) and Logistic Regression (LR) Classifier. | ------ | Construct a corpus consisting of 4k English tweets | Text and Emoticon | Accuracy- 33.43% Precision-0.32 Recall-0.33 F1-0.32 (SGD) Accuracy- 40.10% Precision- 0.37 Recall- 0.40 F1- 0.35 (LR) | During the pre-processing phase, punctuation marks have been eliminated, potentially resulting in the loss of key information. Additionally, the normalization of lengthened words, such as "haapy" and "happy," has not been performed. This lack of normalization can influence the polarity of a tweet. |
| [15], 2019 | Designed a deep neural network called SmileyNet to map images into a low-dimensional emoticon space via noisy and abundant data collected from the web | ------ | Create a unique dataset containing 4 million images gathered from Twitter along with their corresponding emoticons (visual smiley dataset) | Emoticon and Images | Accuracy 3agree- 82.69 4agree- 84.87 5agree- 89.16 | The proposed methodology yields inconspicuous results on the contradiction between visual data and emoticons. Furthermore, the models utilized are not assessed based on their interpretability and explainability. |
| [16], 2019 | Developed a multimodal approach that predicts emoticons by combining textual and visual features. ConvNet was used to retrieve characteristics from the image data, and LSTM was used to extract | Feature-level fusion | Present a multimodal dataset consisting of 15 million tweet having both image-text pairs | Text, Emoticon, and Images | Accuracy Top-1- 20.6 Top-5- 40.3 Top-10- 51.5 Top-100- 89.3 Sample-wise average | Have used the conventional ConvNet instead of testing new pretrained CNN variations like ResNest, ResNext, RegNet, etc. to get more reliable results. |

| Year & Ref. | Methods used for feature extraction | Fusion approach | Dataset utilized | Modality | Experimental results | Limitation |
|---|---|---|---|---|---|---|
| | features from the input text. | | | | precision(msAP)- 27.0 | |
| [17], 2021 | Presented a novel approach that combines Bi-GRU with a self-attention layer (SEER) to get more accurate emoticon predictions for English text. Additionally, the word2vec approach was employed to construct the embedding of the input text. | ------ | NLPCC2014 | Text and Emoticon | Precision- 86.35 Recall- 69.83 F1- 77.22 | Designed for applications that rely on text. Furthermore, better results could be achieved by implementing more advanced models like BERT, RoBERTa, and others. |
| [18], 2021 | Proposed a model titled as CAPER that combines personal and contextual information to provide personalized emoticon recommendations. Various factors such as, user preference, temporal user gender, and text are used to illustrate the different features that may impact a user's selection of emoticons. | ------ | Sina Weibo and Twitter | Text and Emoticon | Precision- 0.135 Recall- 0.524 F1- 0.214 Precision- 0.115 Recall- 0.447 F1- 0.181 | The proposed strategy incorporated a restricted range of features into model. Through our analysis, we discovered that these characteristics do not have much impact on users' selection of emoticons. |
| [19], 2021 | Have analysed early and late fusion methods for integrating emoticons with Arabic text features to enhance the polarity of sentiment. Support vector machines and linear regression models have been used for single modality prediction, while a linear kernel support vector machine is employed for the fusion model. | Hybrid fusion | ASTD (Arabic tweets dataset) | Text and Emoticon | Accuracy- 85.08% Precision- 85.07% Recall- 85.08% F1- 85.08% (Linear kernel support vector machine) | It is possible that the proposed methodology may not be suitable for languages other than Arabic. Furthermore, it's crucial to explore the performance using alternative methods of topic modelling like non-negative matrix factorization, and latent dirichlet allocation. |
| [20], 2022 | Developed a framework which integrates graph attention network (GAT) with encoder transformer network for emoticon prediction, achieving superior performance compared to several baseline models. | ------ | Mu-Emoji | Text and Emoticon | Accuracy- 0.65 F1- 64.27 (single-label) Accuracy- 66.37 F1- 77.53 (multi-label) | Unable to anticipate the emoticons associated to the tweet when sarcasm is expressed. |
| [21], 2022 | Introduced multi-modal technique named as MMCBD to identify cyberbullying by analysing sentiment and emotion. BERT and emoji2vec were used to extract features from text and emoji's, while Bi-GRU was used to maintain | Feature-level fusion | BullySentEmo | Text and Emoticon | Accuracy- 82.87% F1- 82.86% | The model is expensive due to its substantial amount of trainable parameters. |

| Year & Ref. | Methods used for feature extraction | Fusion approach | Dataset utilized | Modality | Experimental results | Limitation |
|---|---|---|---|---|---|---|
| | contextual and semantic information among words. | | | | | |
| [13], 2022 | Suggested a multimodal framework that utilizes Efficient-Net B7 for extracting features from image data and BERT for processing the input text of the tweet. | Feature-level fusion | Multimodal-TwitterEmoticon | Text, Emoticon, and Images | Accuracy- 0.460 F1- 0.461 Acurracy- 0.362 F1-0.354 | Multiple image files in the dataset prepared for the paper are corrupted. The image files require refinement. |
| [22], 2023 | Designed an ensemble-based approach to identify spam in tweets that include emoji's. Also, stated that SVM (Radial Bias Function) has produced the most favourable outcomes. | ------ | SpamID-Pair dataset | Text and Emoticon | Accuracy (Average)- 84.6% F1 (Average)- 77.6% | Have used the conventional approaches instead of validating new pertained variants like BERT, RoBERTa, Alberta etc. to get more reliable results. |
| [23], 2021 | Introduced a model called Seq2Emoji, which utilizes an encoder-decoder based architecture to produce various emoticons for a piece of sentence based on the relationship between sequences, thus ensuring alignment between emoticons and text. The encoder combines ConvNet and Bi-LSTM to improve learning. | ------ | Weibo dataset (Chinese) | Text and Emoticon | Precision- 0.29 Recall- 0.44 F1- 0.28 | The suggested framework might be more robust for only Chinese language than any other widely used language, such as English, Hindi etc. |
| [24], 2022 | This research study introduces MultiEmo, a novel framework to handle multiple tasks. Emoticon prediction serves as primary goal, whereas, emotion recognition is categorized as a secondary task. | ------ | Twitter and GoEmotion Dataset | Text and Emoticon | | To improve accuracy ratings, it is essential to remove numerous non-alphanumeric characters from the Twitter and Go Emotion datasets. |
| [25], 2022 | Introduced an entirely new annotated dataset and examining the effect of location and time on emoticon forecasting. This method utilizes a hybrid BERT-based model that incorporates score-based metrics such as cosine and semantic similarity. | ------ | | Text and Emoticon | Accuracy- 73.32% | The methodology consists of a hybrid architecture that utilizes score-based metrics and, neural networks, specifically semantic and cosine similarity. The research paper does not mention the cosine similarity score, which falls within the range of [0.1,1). |
| [26], 2020 | Proposed a BERT-based solution that surpassed the top baseline approaches in F-score by 2.38% and 7.39%, respectively, for multilingual emoticon prediction. | ------ | SemEval-2018 | Text and Emoticon | F1- 38.52 Precision- 40.64 Recall- 41.76 | The investigation has utilized transfer learning as an innovative component of their strategy. Nevertheless, a multitude of prior |

| Year & Ref. | Methods used for feature extraction | Fusion approach | Dataset utilized | Modality | Experimental results | Limitation |
|---|---|---|---|---|---|---|
| | | | | | | research has previously concentrated on the utilization of transfer learning. |
| [27], 2023 | The purpose of this study to explore the process of classifying emotions through the utilization of facial emoticons. In this study an unsupervised learning techniques has been employed to analyse and extract clues and patterns from real emoticons present in tweets. | ------ | | Text and Emoticon | | An identical method, consistent with previous research, was used to generate the different types of face emoticons from text. This approach also encompasses specific characters that must be eliminated, as they are not utilized in the decision-making process and degrade the accuracy scores. |

[28], [29], [30], [31], [32], [33], [34], [35], [36], and [37] are a few additional recent cutting-edge studies that employ emoji prediction based on sole textual information without the introduction of image modality. Humans possess a natural inclination towards visual stimuli because visuals are an effective means of conveying ideas in a quicker and more efficient manner. However, it is also worth noting that vision plays a significant role in the acquisition of knowledge, accounting for approximately 80% of the information humans learn. As a results, the utilization of images, in conjunction with captions, on diverse social media platforms is steadily growing on a daily basis. **Figure 3** highlights a few motives for incorporating images with captions on social media platforms.

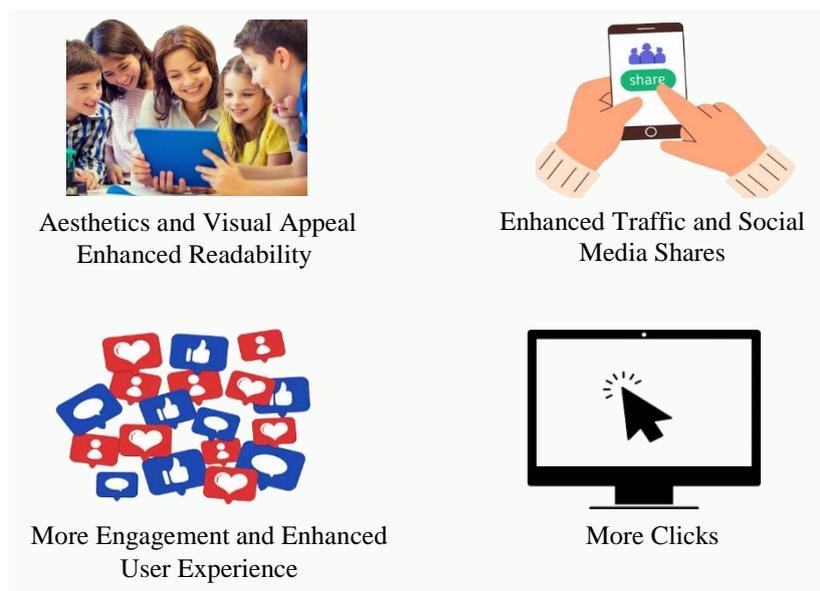

**Figure 3** Few reasons why using visual content in conjunction with captions on social media platforms is advantageous.

The literature survey reveals that the majority of research studies focus on predicting emoticons based solely on textual content. Nevertheless, there exists a substantial need for the prediction of emoticons that considers both image and text. Hence, the main aim of this research study is to showcase the potential enhancement in the accuracy of emoticon prediction by integrating both contextual and visual analysis, as opposed to solely relying on textual analysis by employing advanced state-of-the-art architectures and implementing a more robust multimodal approach.

## 3  Proposed Methodology

This section contains an in-depth analysis of the proposed paradigm for multimodal emoticon predictions. The problem is outlined in the first subsection. Then, a Contrastive Learning based Multimodal Architecture depicted **Figure 4** in consists mainly of 3 components is introduced. The proposed model employs a dual-branch encoder design to accurately map text and images into a common latent space. This functionality is achieved through the joint training of both the encoders. Transformer-based visual encoder, Transformer-based textual encoder and an additional component involves the use of contrastive learning to uncover the hidden relationships within the text and pictures. It has been demonstrated that by integrating the principle of contrastive learning with that of the other two branches yields superior results. For simplicity, emoticon prediction based on text-image analysis is referred to multimodal emoticon prediction.

### 3.1.1  Task Definition

The task of multimodal emoticon prediction is defined as follows: Let $I$ and $T$ denote the sample spaces for an image and text, respectively. An example is comprised of a singular text string accompanied by supplementary visual data. Hence, each example consists of three elements: an image, a piece of text, and an emoticon as a class label. The expression for it is shown below:

$$\mathbb{E} = \{(I^0, T^0, L^0), (I^1, T^1, L^1), \ldots\ldots, (I^i, T^i, L^i), \ldots\ldots, (I^{m-1}, T^{m-1}, L^{m-1})\} \quad (1)$$

where, $\mathbb{E}$ denotes the entire set of instance triplets, $I^j$ symbolizes the images information, $T^j$ represents the text-based data, $L^j$ denotes emoticons numbered from $0 - 9$ as a class label for the $i^{th}$ sample, and $m$ is the count of the total number of examples in the entire dataset.

Multimodal emoticon prediction aims to learn a mapping function $\mathbb{F}: (I^i, T^i) \to L^i$ predict most suitable emoticon for a multimodal tweet $\{(I^i, T^i, L^i) | 0 \le i \le m - 1)\}$. For multimodal emoticon prediction task, $L^i \in \{0,1,2,3,4,5,6,7,8,9\}$, where 0 represents 😭 while 9 denotes 👍.

### 3.2  Contrastive Learning based Multimodal Architecture

We have proposed a multimodal architecture based on the principle of contrastive learning for the emoticon prediction task to effectively simulate the relationship and compatibility between image and text content. **Figure 4** illustrates the proposed architecture. The proposed multimodal architecture comprises of three primary components: An Image encoder, a Text encoder, and a Contrastive learning component. The Image encoder is responsible for acquiring image embeddings, while the Text encoder acquires textual embeddings. The Contrastive learning element examines the pertinent attributes and similarities between the textual and image embeddings obtained in the preceding steps. The proposed model is demonstrated in the form of pseudocode in **Table 2**.

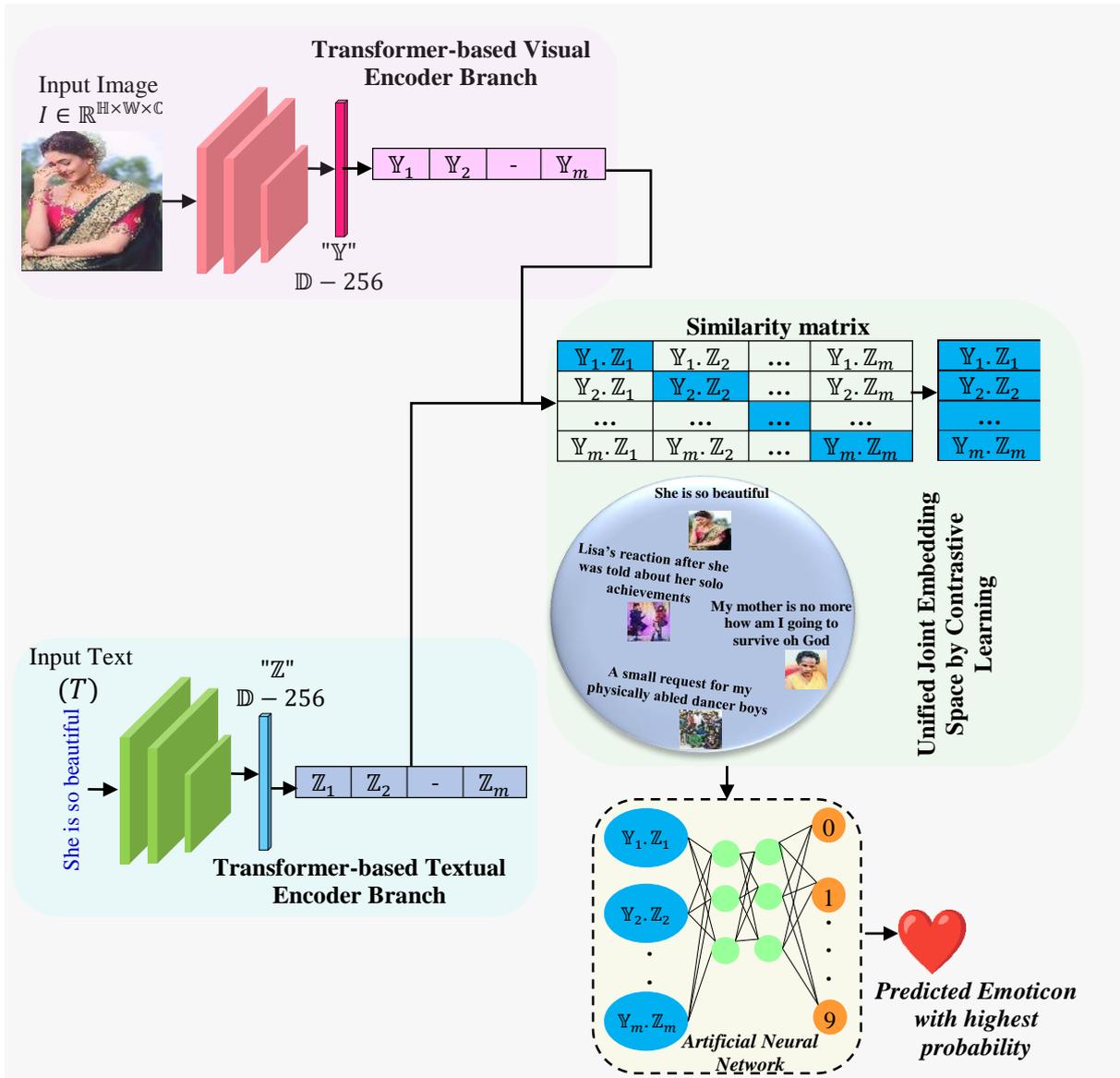

**Figure 4** Proposed Contrastive learning based multi-modal architecture for emoticon prediction.

### 3.2.1 Transformer-based Visual Encoder

This sub-section will provide a comprehensive explanation of the proposed methodology for extracting pertinent details from visual cues. In the realms of image, audio and text analysis, the performance of deep neural network-based architectural designs succeeds over the conventional hand-crafted approaches for classification [38], [39], [40], [41]. The primary reason for their effectiveness lies in their capacity to enhance end-to-end relationships, facilitate autonomous feature learning, ensure efficient scalability, establish semantic representations, and offer flexibility. Furthermore, several researchers have been working to enhance the performance of pre-trained and custom-built ConvNets over the last few years by including attention [42], [43], [44] an additional architectural design component.

The utilization of ConvNets is not deemed essential in modern times. This is because a standalone transformer model [45] can effectively handle visual classification tasks by directly processing sequences of patches of images. Each patch is then converted into a vector via a Large Language Model referred as LLMs and processed using a transformer architecture.

Hence, [45] is capable of gathering and analysing global information in images, unlike ConvNets, which can only extract local aspects.

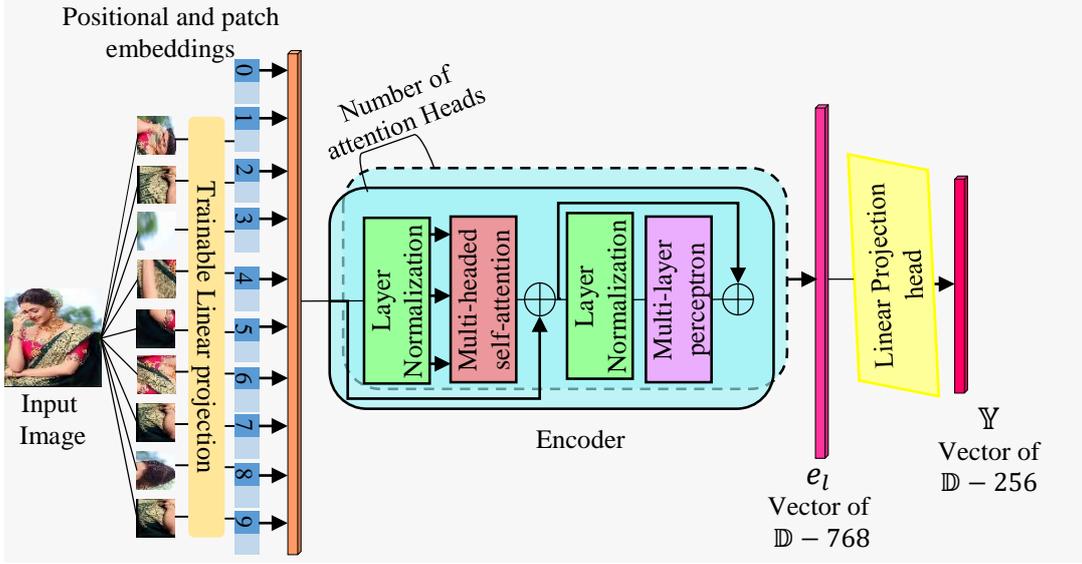

**Figure 5** Image encoder to obtain the embeddings for all the image samples present in the dataset.

Taking these considerations into account, a transformer-based model Vit-base-patch32 depicted in **Figure 5** is utilized to retrieve the most prominent features from the image samples of Multimodal-TwitterEmoticon dataset. Firstly, the initial step involves the scaling of image samples denoted as $I \in \mathbb{R}^{\mathbb{H} \times \mathbb{W} \times \mathbb{C}}$ from the entire collection of the dataset to a resolution of $224 \times 224$. This scaling is done to prepare the samples for subsequent processing. Typically, transformers take a 1-dimensional series of token embeddings as input. We address this by transforming the two-dimensional input image into a sequence of flattened patches $I_p \in \mathbb{R}^{(\mathbb{P}^2 \cdot \mathbb{C}) \times \mathbb{N}}$ of size $(\mathbb{P} \times \mathbb{P}) \in 32 \times 32$. A trainable linear projection is used to transform the acquired patches into $\mathbb{D}$ −dimensional space using **Eq. (1)**. This allows the uniform latent vector size $\mathbb{D}$ of the transformer to be employed throughout all the layers. The projection's end product is called patch embedding. Additionally, to preserve positional information, position embeddings $E_{Position}$ are appended to the patch embeddings. Analogous to BERT's $[class]$ token [46], we augment the patches with a learnable embedding ($e_0^0 = I_{Class}$). The final obtained patches are then pass through a visual encoder layer which is nothing but the $h$ number of self-attention operations, called "heads", which will run in parallel. The final outcome of the Transformer-based visual encoder is used as the representation for the images denoted as $e_h$ of $\mathbb{D} − 768$ in **Eq. (2)** and **Eq. (3)**. The GELU non-linearity is used on two consecutive layers to build the multilayer perceptron.

Finally, we need to translate the image and text embeddings into the same vector space so that we may process this acquired vector together with the textual embedding. This is why the obtained vector of $\mathbb{D} − 768$ goes through a linear projection head, basically just an artificial neural network with a linear activation function to get the desired result. Thus, the final vector $\mathbb{Y}$ will be $\mathbb{D} − 256$, as stated in **Eq. (4)**, and will be used for further operations.

$$e_0 = [I_{Class}; I_p^1 E; I_p^1 E; \ldots \ldots; I_p^1 E] + E_{Position} \tag{1}$$

$$e_h' = Multi-headed\ self\ attention\bigl(Layer\ Normaliztion(e_{h-1})\bigr) + e_{h-1} \tag{2}$$

$$e_h = Multi - layer\ perceptron\big((e'_h)\big) + e'_h \tag{3}$$

$$\mathbb{Y} = Linear\ projection\ head(e_h^0) \tag{4}$$

where, $\mathbb{H} \times \mathbb{W}$ denotes the height and width of the original image sample $I$, $\mathbb{C}$ symbolizes the depth of an image, $\mathbb{P} \times \mathbb{P}$ denotes the resolution for each transformed patch $I_p$ of an image, $h = \{1,2,...,H\}$ denotes number of attention heads, $E \in \mathbb{R}^{\mathbb{D} \times (\mathbb{P}^2 \cdot \mathbb{C})}$ and $E_{Position} \in \mathbb{R}^{\mathbb{D} \times (\mathbb{N}+1)}$.

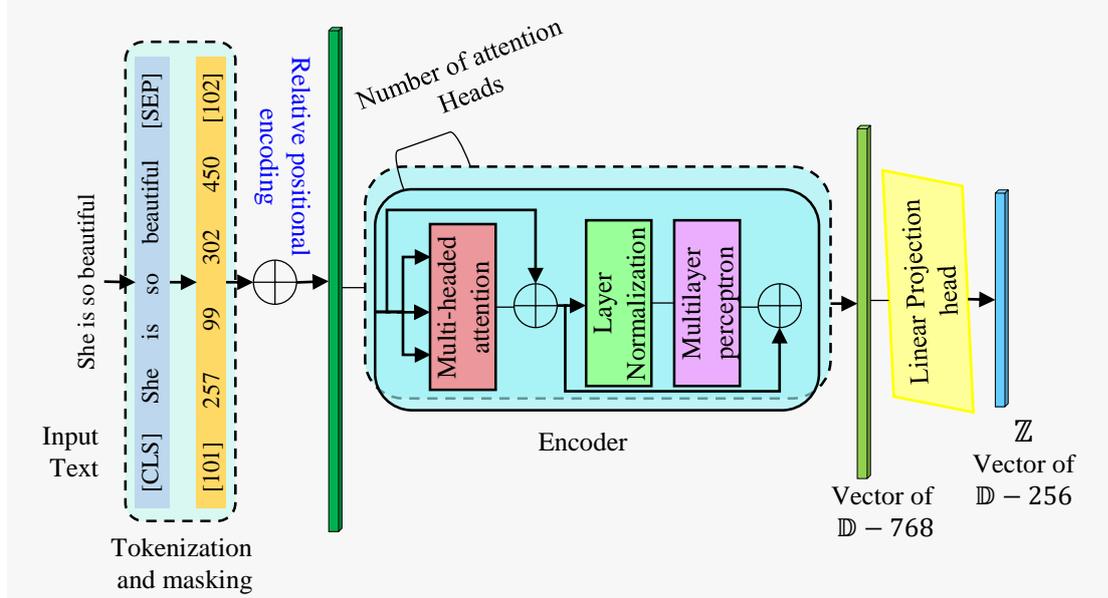

**Figure 6** Text encoder to obtain the embeddings for all the text samples present in the dataset.

### 3.2.2 Transformer-based Textual Encoder

The latest advancements in transformer-based models incorporate a self-attention mechanism to prioritise relevant information while disregarding irrelevant information. Recent studies primarily utilised these architectures [46] exclusively for text processing. Although there are different variants of BERT that aim to extract text features, but they differ from the [47] architecture design due to their reliance on absolute position encoding rather than relative position codification [48]. The [47] utilises relative positional encoding to incorporate information between pairwise positions, whereas absolute positional encoding does not take this into consideration. In absolute positional encoding, the embeddings for each location are initialised at random. As a result, the relationship between different positions is unknown. Instead of random embedding initialization, relative positional encoding generates a pairwise vector of size $(V, 2 * V - 1)$, with the row index representing the desired word and the column index representing its position distance from previous and subsequent words. This information regarding relative positioning is dynamically integrated into the keys and values as part of the computation process in attention modules. Therefore, it is advantageous to use a transformer-based model that supports relative positional encoding. This feature provides greater flexibility to the model and leads to more accurate result.

Considering these factors, we have employed the encoder of [47] to transform the text into their respective embedding's. During the first step, the sentence $T$ provided as input to the Transformer-based text encoder is partitioned into tokens $T = \{t_1, t_2, ......, t_n\}$, and each token is subsequently transformed into a vector representation $T^{vectorization} = \{t_1^v, t_2^v, ......, t_n^v\}$. Next, the acquired vectors are sent for relative position encoding. The results achieved through relative position encoding are then fed into the transformer-based text encoder component, which produces a text embedding vector $\mathbb{T}$ of size $\mathbb{D} - 768$. In order to ensure that the image and text embedding's are aligned in the identical vector space, the $\mathbb{D} - 768$ is sent to a linear projection head to compress the text embedding to a vector $\mathbb{Z}$ of size $\mathbb{D} - 256$.

### 3.2.3 Contrastive Learning

The primary goal of multi-modal learning is to understand the relationships between images and text in a given batch $B = \{I^i, T^i\}_{i=1}^m$, which consists of $m$ examples represented as ($T^i$, $T^i$). Motivated by the architectural design of Contrastive visual-textual pre-training, we applied the concept of a similarity matrix to examine the relevant features and similarities between the embeddings of image-text pairs acquired by the transformer-based encoder in previous stages. In order to do this, firstly contrastive visual-textual pre-training model [49] trains an image encoder $f(I^i, L^i)$ and a text encoder $g(T^i, L^i)$, such that the embeddings of image-text pairings $\{I^i, T^i\}_{i=1}^m = \{f(I^i, L^i), g(T^i, L^i)\}_{i=1}^M$ become more similar to each other. It is important to mention that $\mathbb{Y}$ and $\mathbb{Z}$ are unit vectors of size $\mathbb{D} - 256$ that have been normalised using the $\mathbb{L}_2$ norm in the encoding phase. These vectors are then located on the similar hypersphere. Contrastive visual-textual pre-training model employs the loss function $C(\cdot,\cdot)$ as specified in the cited approach [13] with the goal of ensuring that pairs ($\mathbb{Y}$, $\mathbb{Z}$) possess both similarity and distance. The formulation is represented by the **Eq. (5) and Eq. (6)**:

$$C(I,T) = \frac{1}{m}\sum_{i=1}^m -log\frac{\exp(sim(\mathbb{Y}_i,\mathbb{Z}_i)/\tau}{y\sum_{i=1}^m \exp(sim(\mathbb{Y}_i,\mathbb{Z}_i)/\tau} \quad (5)$$

$$\mathcal{L}_{\text{Contrastive visual-textual}} = \frac{1}{2}(C(\mathbb{Y}_i,\mathbb{Z}_i) + C\,\mathbb{Y}_i,\mathbb{Z}_i) \quad (6)$$

Similar to other approaches in computational linguistics, [49] uses a dot product to calculate similarity ($sim(.,..)$) between two vectors. It also employs a learnable temperature parameter ($\tau$) to adjust the magnitude of the observed similarity and $\mathcal{L}_{\text{Contrastive visual-textual}}$ is the mean loss of combined image and text pairs. In general, the model learns a multi-modal embedding space by training its encoders to maximize the cosine similarity between the picture and text embeddings of the $N$ correct pairs in the batch, while minimizing the cosine similarity between the embeddings of the $N^2 - N$ incorrect pairings. Hence, by representing both images and texts using the coherent embedding space, our proposed contrastive learning-based model is capable of doing two essential tasks: ($a$) Optimize the cosine similarity between the image and text embeddings for $N$ real pairs in the batch, aiming to maximize it. ($b$) Additionally, minimize the cosine similarity between the embeddings of $N(N-1)$ erroneous pairings. During the pre-training phase of [49], the model is trained using a contrastive loss function, as depicted in **Eq. (6).** The main objective of this loss function is to promote the aggregation of embeddings for related or *positive pairings* (text and picture that match) while simultaneously driving apart the embeddings for unrelated or *negative pairs* (text and image that do not match). The purpose of the model is to minimise the loss for positive pairs and maximise the loss for negative pairs. The pair with the maximum cosine similarity scores, as indicated by **Eq. (7),** is between two $\mathbb{D}$-*dimensional* vectors, referred to as $\mathbb{Y}$ and $\mathbb{Z}$. These obtained vectors are then fed into the simple artificial neural network and processed using softmax as an activation function to provide probabilities for each label.

$$Sim(I,T) = \delta = \frac{\mathbb{Y} \cdot \mathbb{Z}}{||\mathbb{Y}|| \cdot ||\mathbb{Z}||} = \frac{\Sigma_{i=1}^{m} \mathbb{Y}_i \times \mathbb{Z}_i}{\sqrt{\sum_{i=1}^{m}(\mathbb{Y}_i)^2} \times \sqrt{\sum_{i=1}^{m}(\mathbb{Z}_i)^2}} \qquad (7)$$

**Table 2** Pseudocode for the proposed Contrastive Learning-Based Multimodal Architecture.

| |
|---|
| **Pseudocode:** Emoticon prediction using Contrastive Learning-Based Multimodal Architecture |
| **Objective:** To acquire knowledge of a mapping function $\mathbb{F}: (I^i, T^i) \to L^i$ from a set of multimodal tweets $\{(I^i, T^i, L^i \mid 0 \leq i \leq m-1)\}$. <br> **Input:** Image set $I = \{I^0, I^1, ..., I^i, ..., I^{m-1}\}$ and Text set $T = \{T^0, T^1, ..., T^i, ..., T^{m-1}\}$, $m$ is the count of the total number of examples in the entire dataset. <br> **Output:** Multimodal emoticon prediction task, $L^i \in \{0, 1, 2, 3, 4, 5, 6, 7, 8, 9\}$, where **0** represents 😭 while **9** denotes 👍. |
| 1. Extract image features "$e_h$" of size "$\mathbb{D} - 768$" from the visual samples using Transformer-based visual encoder by Eq. (1), Eq. (2), and Eq. (3); <br> 2. Extract text features "$\mathbb{T}$" of size "$\mathbb{D} - 768$" from the caption samples using Transformer-based textual encoder. <br> 3. for $E_{epochs} \leftarrow 1$ to Epochs do; <br><br> #identify and extract modality-specific feature representations <br> $e_h$ = *visual_encoder(I)* **Eq.(1), Eq.(2), and Eq.(3)**; <br> $\mathbb{T}$ = *textual_encoder(T)*; <br><br> # joint multimodal embedding where $W_I$ and $W_T$ are weights which are learned projection of image and text to embedding <br> $\mathbb{Y} = \mathbb{L}_2\_normalization(np.dot(e_h, W_I), axis = 1)$ <br> $\mathbb{Z} = \mathbb{L}_2\_normalization(np.dot(\mathbb{T}, W_T), axis = 1)$ <br><br> # cosine similarity between pairs of vectors <br> $logits = np.dot(\mathbb{Y}, \mathbb{Z}.T) * np.exp(\tau)$ <br><br> # calculating contrastive loss function to optimize the cosine similarity between the text and image embedding for $N$ real pairs in the batch <br> $labels = np.arange(m)$ <br> $loss\_I = cross\_entropy\_loss(logits, labels, axis = 0)$ <br> $loss\_T = cross\_entropy\_loss(logits, labels, axis = 1)$ <br> $loss = (loss\_I + loss\_T)/2$ *using* **Eq.(5), Eq.(6), and Eq.(7)** <br><br> # At last The pair with the maximum cosine similarity scores are then fed into the simple artificial neural network to provide probabilities for each label. <br> $L^i = Artificial\ neural\ network((\mathbb{Y}, \mathbb{Z})_{Maximum\_cosine})$ find the loss and execute the backpropagation; <br><br> end |

## 4 Experimental Setup and Results

The following section of the research article will present a detailed description of the optimal experimental configurations, the dataset utilised, and the experimental results obtained using our proposed approach.

### 4.1 Experimental Configuration

It is crucial to determine an appropriate range for hyper-parameters in a learning algorithm, as they play a significant role in controlling the learning process. The hyper-parameters underwent a random testing process using various values. As a result, the values that yielded the most optimal outcomes for our proposed framework were subsequently set as fixed. Since, the joint embedding from both modalities was obtained using the concept of contrastive learning. Therefore, the hyper-parameter values used were almost similar to those in [49]. The encoders utilised for both images and text in our proposed model referred as Contrastive Learning based Multimodal Architecture are equipped with 12 number of attention heads. The image samples denoted as $I \in \mathbb{R}^{\mathbb{H} \times \mathbb{W} \times \mathbb{C}}$ from the dataset are scaled to a resolution of $224 \times 224$ before being passed to a visual encoder. Adam is used as an optimizer with the default learning rate of 0.001. The values of $\beta_1$ and $\beta_2$ for using Adam are set to 0.9 and 0.99. A dropout rate of 0.2 has been implemented. The contrastive loss function is also used (shown in **Equation (5)**) with a temperature scaling factor $\tau = 0.1$ and margin $M = 0.7$ to combine the embeddings of images and text with the highest similarity score. The proposed architecture is fine-tuned for 20 epochs with a batch size of 32. By the 20th epoch, it becomes evident that there has been no noticeable rise in accuracy, as the results reach a point of saturation. From a total of 21k samples, 16k were allocated for training purposes, while the remaining 5k samples were dedicated to testing our proposed model.

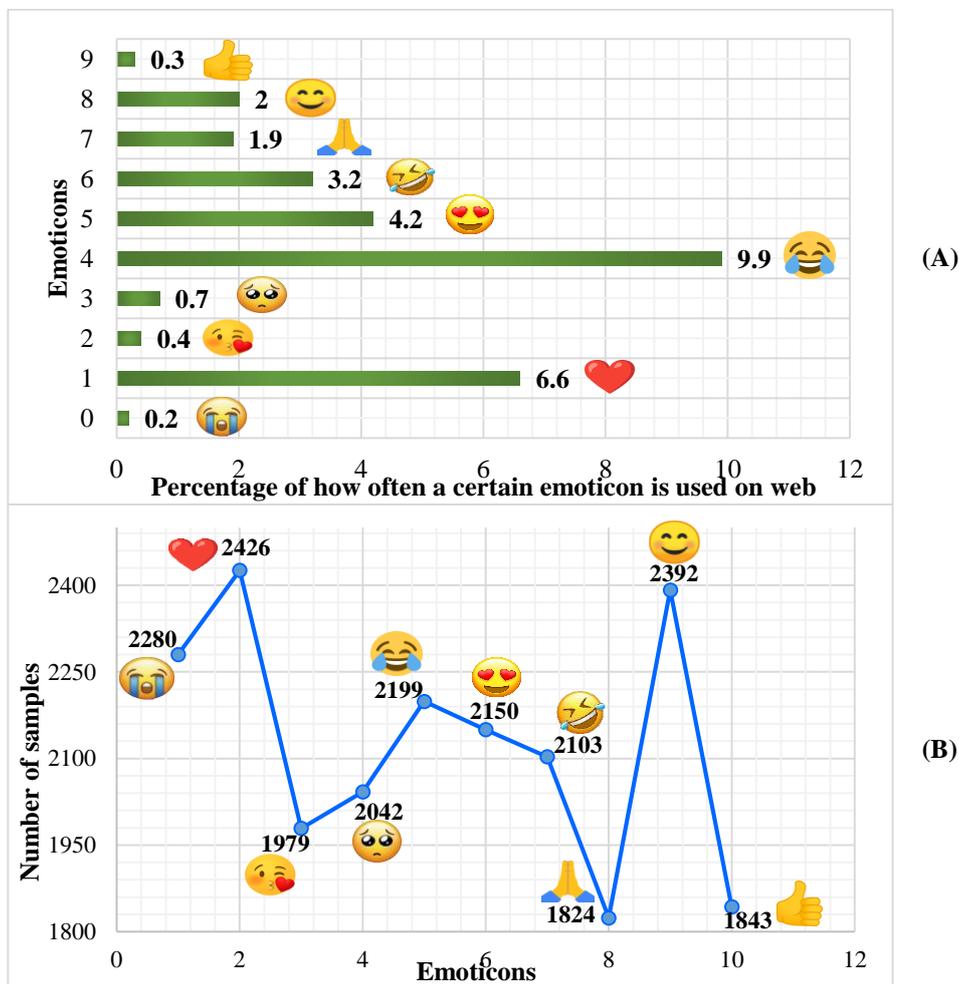

**Figure 7 (A)** provides information about how often multiple emoticons are utilized on web, while **7 (B)** indicates the number of samples corresponds to each emoticons in Multimodal-TwitterEmoticon dataset.

## 4.2 Hardware Configuration

An Azure virtual machine with premium specifications was used for training and evaluation of the proposed model. This machine provides 100 GB of hard drive space, NVIDIA-A100 GPU with 80 GB of graphics memory, CUDA version 12.4, and 384 GB of RAM. We utilised an $8 \times A100$ GPU cluster for 41 days, which is roughly comparable to 1000 GPU hours, and a pool memory of 640 GB to train the entire model from end to end. To derive useful information from text-image pairings, models were constructed using the Pytorch frameworks.

### 4.3 Dataset Description

We have employed the Multimodal-TwitterEmoticon dataset, provided by Ebrahimian et al. [13], to evaluate our suggested approach for emoticon prediction. Multimodal-TwitterEmoticon dataset consists of 21k English tweets, each containing an image, the accompanying text, and a single emoticon. Emoticons used in tweets varies based on their popularity. **Figure 7(A)** illustrates the frequency in percentage for most widely used emoticons. These ten most prevalent and often used emoticons have been taken into account for analyses. **Figure 7(B)** also depicts the statistical information regarding the number of samples associated with each 10 emoticons across the entire dataset where 0 represents 😭 while 9 denotes 👍. The experiment focuses on predicting emoticons based on image-text pairs. From a total of 21k samples, 16k were allocated for training purposes, while the remaining 5k samples were dedicated to testing our proposed model.

### 4.4 Experimental Results, Baseline Comparison and Analysis

We evaluate our proposed Contrastive Learning-Based Multimodal Architecture on Multimodal-TwitterEmoticon dataset provided by provided by Ebrahimian et al. [13]. The experimental results of our model are presented in **Table 3**. In addition to calculating the overall accuracy, macro-average and weighted-average, we have computed the $Precision$, $Recall$, and $F1-score$ for each class separately to better analyse the results. By taking into consideration every aspect of the confusion matrix represented in **Figure 10**, Matthews Correlation Coefficient ($MCC-Score$) provides a more fair evaluation. Also, when the repercussions of false positives and false negatives differ, MCC becomes more relevant than F1. Hence, we have also calculated MCC-Score.

**Figure 8** shows the training and validation loss curves, which can be used to better rely on the results. Receiver operating characteristic (ROC) curve is considered to be more significant when evaluating model's performance. Unlike $Accuracy$, which solely evaluates the number of right predictions, this statistic takes into account the trade-offs between $Recall$ and $Precision$. It reveals the degree to which the model can differentiate across categories. A higher AUC indicates that the model is effective at making accurate predictions. **Figure 9** displays the ROC curve, which has an area under the curve (AUC) of **0.82**.

**Table 3** Experimental Results on Multimodal-TwitterEmoticon dataset for emoticon prediction.

| $Class\ Labels$ | $Precision$ | $Recall$ | $F1-score$ |
|---|---|---|---|
| 0 | 0.85 | 0.83 | 0.84 |
| 1 | 0.92 | 0.94 | 0.93 |
| 2 | 0.89 | 0.91 | 0.90 |
| 3 | 0.93 | 0.92 | 0.92 |
| 4 | 0.90 | 0.93 | 0.91 |
| 5 | 0.88 | 0.90 | 0.89 |
| 6 | 0.91 | 0.89 | 0.90 |
| 7 | 0.94 | 0.95 | 0.94 |
| 8 | 0.95 | 0.93 | 0.94 |
| 9 | 0.93 | 0.91 | 0.92 |
| $Macro-average$ | **0.91** | **0.91** | **0.91** |

| Class Labels | Precision | Recall | F1 − score |
|---|---|---|---|
| Weighted − average | 0.91 | 0.91 | 0.91 |
| Overall Accuracy | 91.00% | | |
| MCC − Score | 0.90 | | |

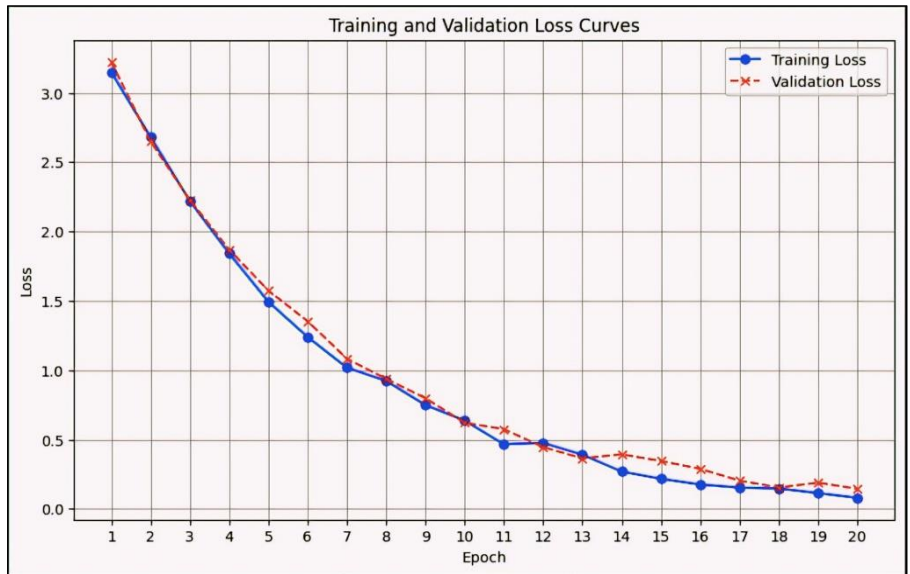
**Figure 8** Training and validation loss curves for the proposed architecture.

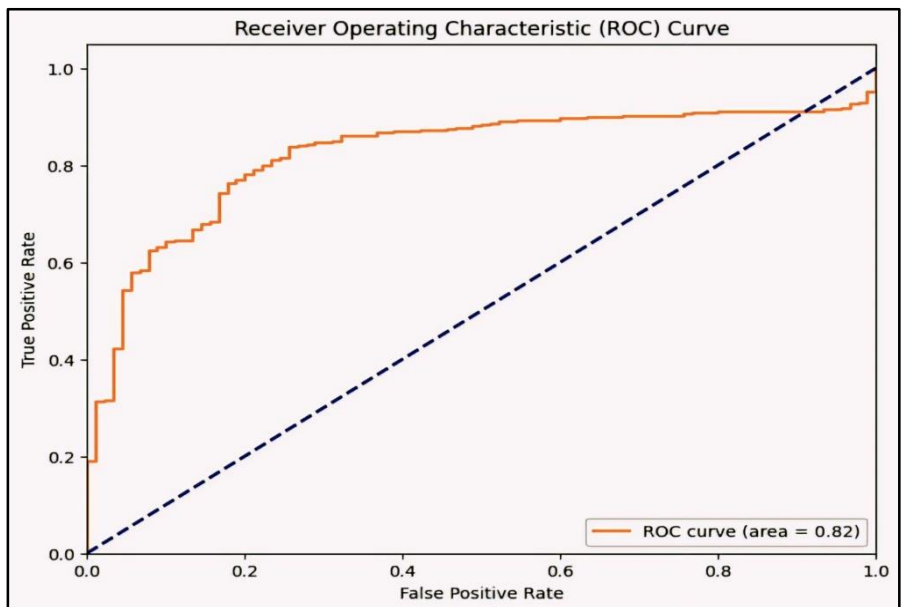
**Figure 9** Receiver operating characteristic curve for the proposed approach.

After examining several recent cutting-edge research articles in the field of emoticon prediction, it becomes clear that there is a scarcity of studies that have effectively combined visual and textual data to predict emoticons. As a result, to assess the strength of our proposed model, we have identified only two baseline methods that incorporate both text as well as images to predict emoticons.

In baseline 1 Francesco et al. [32] scraped Instagram posts to collect multimodal content for emoticon prediction. This dataset is not publically accessible. In this work, a Bi-LSTM model [50] as well as FastText [51] was utilised to extract the most significant features from the text samples present in the dataset. Additionally, a ResNet-101 model was employed to obtain features from the picture samples of the dataset. Baseline 1 also includes a comparison of the FastText model with the character and word based B-LSTMs proposed by Barbieri et al. [12]. The FastText model proves to be highly effective, even outperforming the character-based B-LSTM in emoticon forecasting task. The findings of this research demonstrate that FastText works well for representing brief text from social networking platforms like Instagram or Twitter. Due to the unavailability of the dataset for this study, we have utilised the Multimodal-TwitterEmoticon dataset provided by Ebrahimian et al. [13] to evaluate our approach and compare it with the method proposed by Francesco et al. [32]. The results for the evidence are presented in **Table 4**.

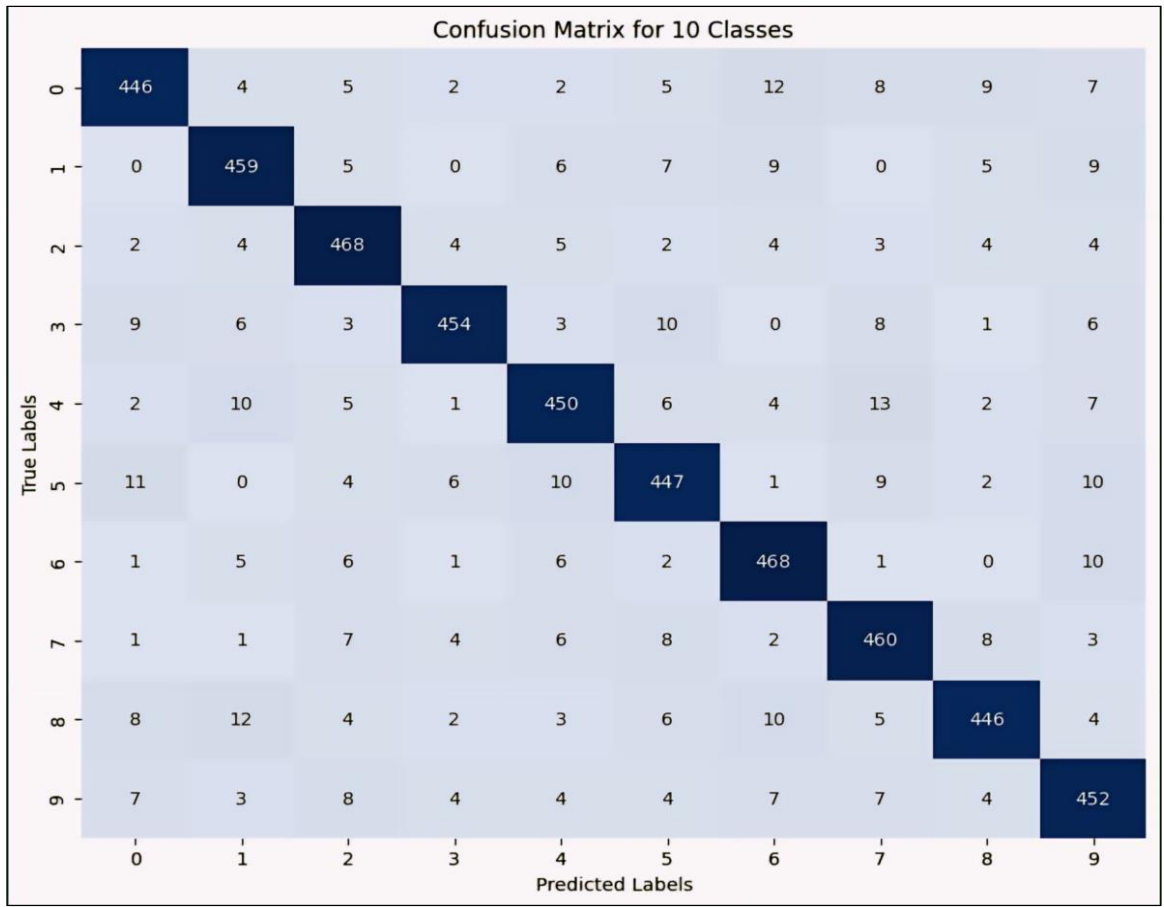

**Figure 10** Confusion matrix obtained for the test samples for the ten class labels by using our proposed method.

**Table 4** Experimental Results on Multimodal-TwitterEmoticon dataset by using the method of baseline 1 (Francesco et al. [32]).

| $Class\ Labels$ | $Precision$ | $Recall$ | $F1-score$ |
|---|---|---|---|
| 0 | 0.75 | 0.72 | 0.73 |
| 1 | 0.72 | 0.73 | 0.72 |
| 2 | 0.69 | 0.70 | 0.69 |
| 3 | 0.73 | 0.72 | 0.72 |
| 4 | 0.70 | 0.73 | 0.71 |
| 5 | 0.78 | 0.69 | 0.73 |
| 6 | 0.71 | 0.70 | 0.70 |
| 7 | 0.73 | 0.74 | 0.73 |
| 8 | 0.74 | 0.73 | 0.73 |
| 9 | 0.74 | 0.71 | 0.72 |
| $Macro-average$ | **0.73** | **0.72** | **0.72** |

| Class Labels | Precision | Recall | F1 − score |
|---|---|---|---|
| Weighted − average | 0.73 | 0.72 | 0.72 |
| Overall Accuracy | 73.00% | | |
| MCC − Score | 0.71 | | |

In baseline 2, Ebrahimian et al. [13] also scrapped multimodal post from the Twitter platform. In this research EfficientNet-B7 has been utilized to obtain visual features. In addition, a topic modelling technique named Latent Dirichlet Allocation is also employed to discover embedded topics within the text. The extracted topics are then combined with the transformer-based network to enhance the efficacy of the proposed model. As compared to Ebrahimian et al. [13] our proposed Contrastive Learning-Based Multimodal Architecture performs far better in terms of multiple performance metrics. **Table 5** highlights the comparison of our results with both the baseline approaches discussed above. A comparative evaluation of the baseline methods in terms of $Accuracy$, $Precision$, $Recall$, and $F1 − score$ with our suggested framework on Multimodal-TwitterEmoticon dataset is illustrated in graphical form in **Figure 11.**

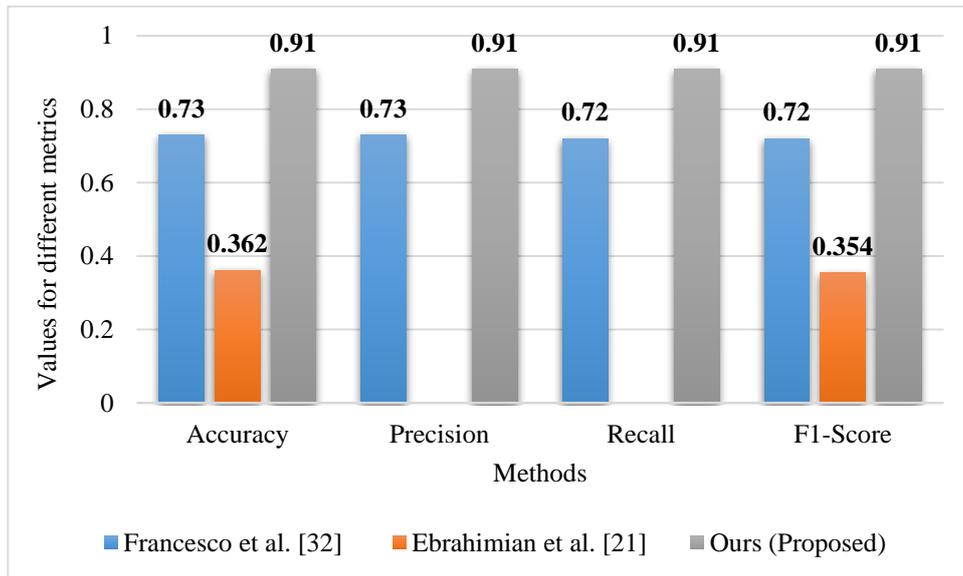

**Figure 11** Graphical representation to represent that our proposed model surpasses existing cutting-edge baselines.

**Table 5** Comparison of our proposed method with baseline methods on Multimodal-TwitterEmoticon dataset.

| Methods | Accuracy | Precision | Recall | F1 − score |
|---|---|---|---|---|
| Francesco et al. [32] | 0.73 | 0.73 | 0.72 | 0.72 |
| Ebrahimian et al. [13] | 0.362 | - | - | 0.354 |
| **Ours (Proposed)** | **0.91** | **0.91** | **0.91** | **0.91** |

Additionally, we have evaluated the predictions in the absence of contrastive learning to further test the robustness of our suggested model. In brief, we have combined the information retrieved from the image and text encoders and then passed them through a basic artificial neural network without depending on contrastive learning approach. However, it significantly diminishes the outcomes. Furthermore, we have evaluated the efficacy of our suggested model by means of t-Distributed Stochastic Neighbour Embedding (t-SNE). **Figure 12** shows that the model can create distinct clusters for different classes. Accordingly, it is able to glean valuable insights from the data.

Hence, based on the obtained results, the following conclusion can be deduced. The proposed architecture demonstrates superior results compared to the baseline approaches, as it outperforms them in terms of $Accuracy$, $Precision$, $Recall$, and $F1-score$. Thus it is reasonable to conclude that our contrastive-based multimodal architecture works effectively by uncovering the hidden relationships within the text and images. In general, the model learns a multi-modal embedding space by the joint training of two encoders to maximize the cosine similarity between the picture and text embedding of the correct pairs in the batch, while minimizing the cosine similarity between the embedding of the incorrect pairings.

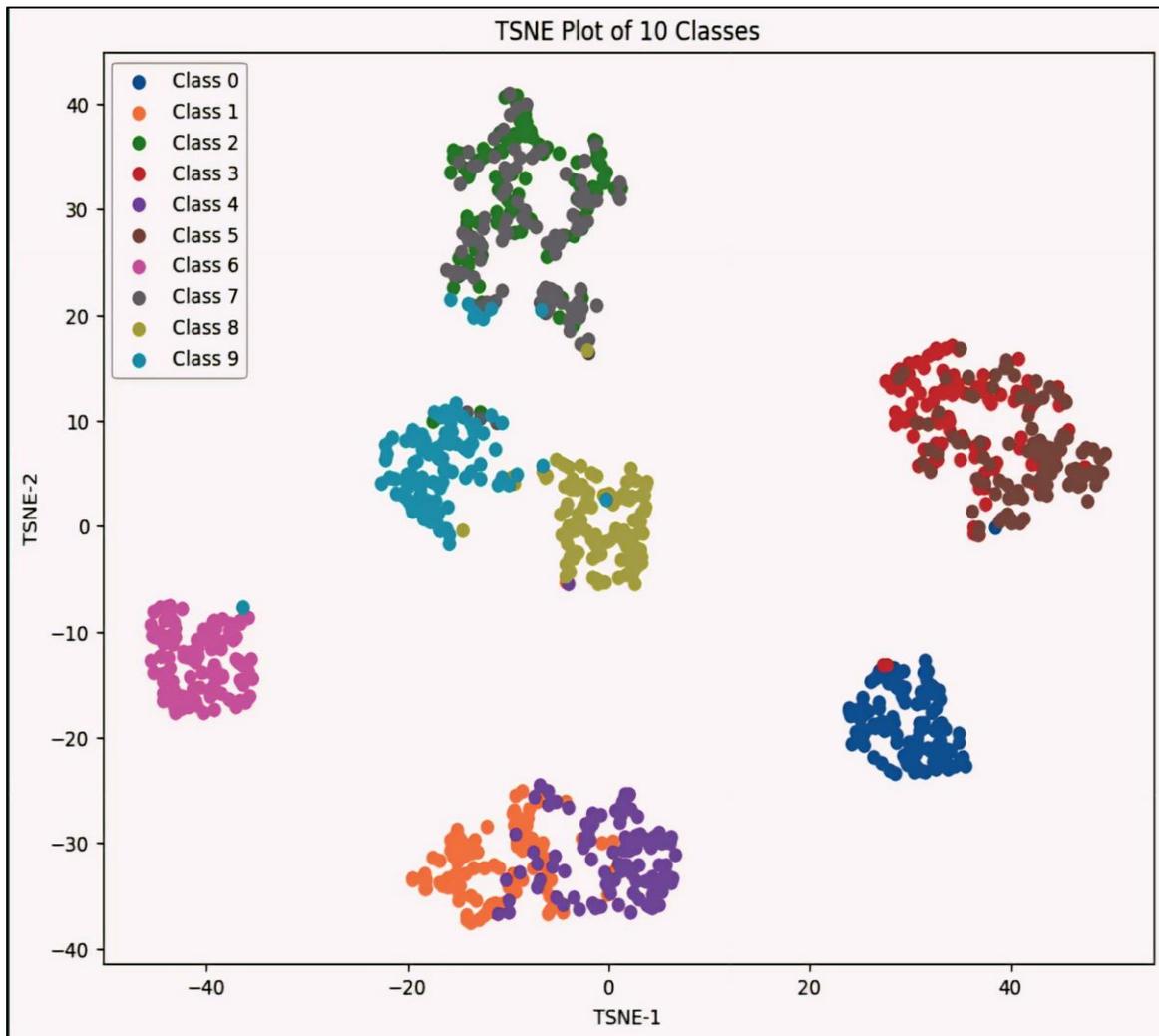

**Figure 12** t-Distributed Stochastic Neighbour Embedding representation to test the strength of the proposed architecture.

### 4.5 Ablation Study

To evaluate the efficacy of each individual element in the proposed architecture, we have performed an ablation study. In the ablation experiment, the text encoder proposed in the study is substituted with BERT-base, BERT-large[52], Roberta[53] and T5 models, while the image encoder is replaced with the ResNet-101[54], EfficientNet-B7[55], ResNext[56], RegNet[57] and Vit-base-patch32 models. The results have undergone analysis using both unimodal and multimodal configurations.

**Table 6** Ablation study to assess the robustness of the text branch for unimodal configuration.

| Text Encoder | *Accuracy* | *Precision* | *Recall* | *F1 − score* |
|---|---|---|---|---|
| BERT-base [52] | 70.06 | 71.64 | 71.42 | 71.53 |
| BERT-large [52] | 75.98 | 75.66 | 74.90 | 75.28 |
| Roberta[53] | 77.63 | 77.34 | 77.15 | 77.24 |
| T5 [47] | **81.42** | **81.83** | **80.97** | **81.40** |

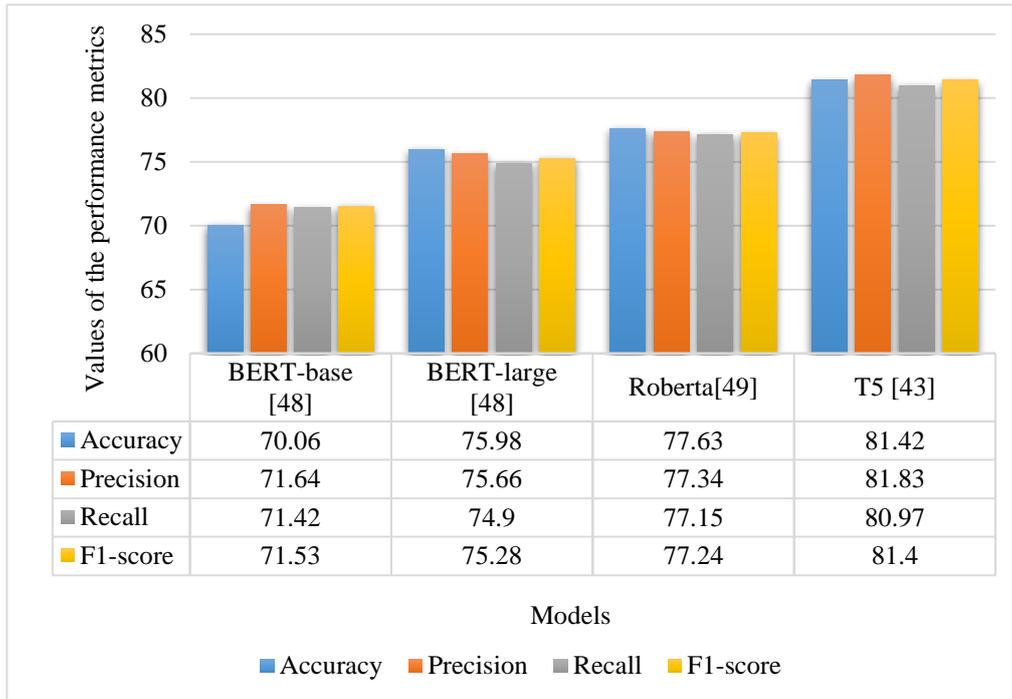

**Figure 13** A graphical representation for an ablation study that aims to evaluate the resilience of the text branch in a unimodal configuration.

Based on the results of the ablation study presented in **Table 6** and **Figure 13** it is clear that the architecture used for the text encoder in the proposed framework is more efficient. The reason behind this is that the model such as BERT-base [52], BERT-large[52], and Roberta[53] supports absolute positional encoding instead of relative positional encoding. As a result, the relationship between different positions is unknown. Since, T5 [47] utilises relative positional encoding to incorporate information between pairwise positions. This information regarding relative positioning is dynamically integrated into the keys and values as part of the computation process in attention modules. Therefore, it is advantageous to use a transformer-based model that supports relative positional encoding. This feature provides greater flexibility to the model and leads to more accurate result.

**Table 7** Ablation study to assess the robustness of the image branch for unimodal configuration.

| Image Encoder | *Accuracy* | *Precision* | *Recall* | *F1 − score* |
|---|---|---|---|---|
| ResNet-101[54] | 72.10 | 72.34 | 72.30 | 72.31 |
| EfficientNet-B7[55] | 72.24 | 71.60 | 71.97 | 71.68 |

| Image Encoder | *Accuracy* | *Precision* | *Recall* | *F1 − score* |
|---|---|---|---|---|
| ResNext[56] | 69.82 | 68.97 | 68.64 | 68.70 |
| RegNet[57] | 67.63 | 68.41 | 68.36 | 68.39 |
| Vit-base-patch32 [45] | **84.75** | **84.61** | **84.43** | **84.52** |

For the visual encoder branch of the proposed architecture, similar to the textual branch, various types of image encoder have been tested. Furthermore, from the experimentation of the ablation study depicted in

**Table 7** and **Figure 14** it has been discovered that [41] outperforms at extracting the most salient elements from visual data. This is because a standalone transformer model [45] can effectively handle visual classification tasks by directly processing sequences of patches of images. Each patch is then converted into a vector via a Large Language Model referred as LLMs and processed using a transformer architecture. Hence, [45] is capable of gathering and analysing global information in images, unlike ConvNets, which can only extract local aspects. Also, [45] can be regarded as superior to different variants of ConvNet architectures, particularly in terms of scalability and the attention mechanism.

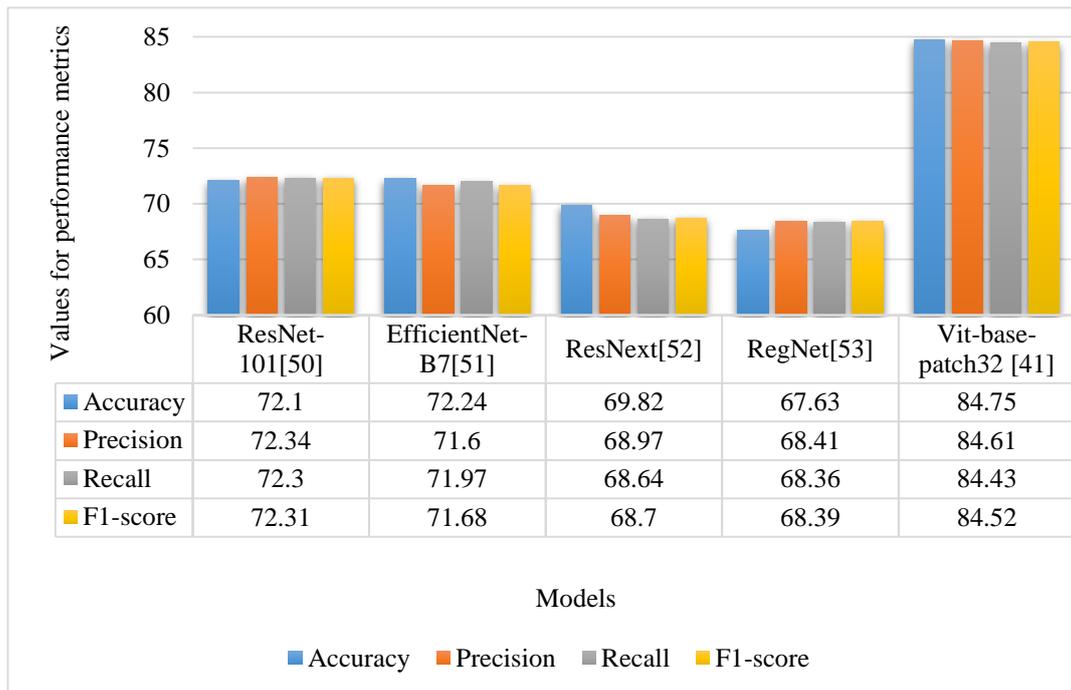

**Figure 14** A graphical representation for an ablation study that aims to evaluate the resilience of the image branch in a unimodal configuration.

Based on the experimental results of the ablation study, it has been determined that the configurations [47] and [45] yield the most optimal outcomes for the unimodal setup. Therefore, we have employed the fusion of these encoders in the proposed contrastive-based architecture for the ultimate prediction.

## 5 Conclusion and Future Directions

The main aim of this research is to introduce an innovative and novel framework called Contrastive Learning-Based Multimodal Architecture for the prediction of emoticons. This architecture is designed to work effectively with image-text pairs. Our suggested approach surpasses existing cutting-edge techniques in analysing the interplay between caption and visual modalities. The sentiment elicited by a phrase can exhibit variability in various scenarios, depending upon the context. Thus, it is essential to employ a blend of textual and visual information to attain more accurate prediction. Motivated by this, we developed a novel dual-branch architecture comprises of three primary components: Transformer-based visual encoder, Transformer-based textual encoder and an additional component involves the use of contrastive learning to uncover the hidden relationships within the text and images. In general, the model learns a multi-modal embedding space by the joint training of two encoders to maximize the cosine similarity between the picture and text embeddings of the $N$ correct pairs in the batch, while minimizing the cosine similarity between the embeddings of the $N^2 - N$ incorrect pairings. A thorough analysis on one of the standard datasets referred as Multimodal-TwitterEmoticon shown that our suggested strategy outperforms strong baseline models.

Despite the excellent outcomes that have been achieved, there remain numerous opportunities for research. These includes creation of publically accessible dataset with an objective of mutli-label emoticon prediction, enhancing feature extraction methods, integrating adversarial learning capabilities to the fusion module, investigating a broader range of multimedia data, encompassing both video and acoustic formats, in order to potentially uncover more detailed semantic relations.

## References

[1] M. Kejriwal, Q. Wang, H. Li, and L. Wang, "An empirical study of emoji usage on Twitter in linguistic and national contexts," *Online Social Networks and Media*, vol. 24, p. 100149, Jul. 2021, doi: 10.1016/j.osnem.2021.100149.

[2] S. Aoki and O. Uchida, "A method for automatically generating the emotional vectors of emoticons using weblog articles," in *Proceedings of the 10th WSEAS International Conference on Applied Computer and Applied Computational science*, ACM, Association for Computing Machinery, 2011, pp. 132–136. Accessed: Apr. 08, 2024. [Online]. Available: https://dl.acm.org/doi/10.5555/1965610.1965632

[3] F. Barbieri, F. Ronzano, and H. Saggion, "What does this Emoji Mean? A Vector Space Skip-Gram Model for Twitter Emojis," in *Proceedings of the Tenth International Conference on Language Resources and Evaluation (LREC'16)*, European Language Resources Association (ELRA), May 2016, pp. 3967–3972. Accessed: Apr. 08, 2024. [Online]. Available: https://aclanthology.org/L16-1626

[4] F. Barbieri, G. Kruszewski, F. Ronzano, and H. Saggion, "How Cosmopolitan Are Emojis? Exploring Emojis Usage and Meaning over Different Languages with Distributional Semantics," in *Proceedings of the 24th ACM international conference on Multimedia*, ACM, Association for Computing Machinery, pp. 531–535. doi: https://doi.org/10.1145/2964284.2967278.

[5] B. Eisner, T. Rocktäschel, I. Augenstein, M. Bošnjak, and S. Riedel, "emoji2vec: Learning Emoji Representations from their Description," in *Proceedings of the Fourth International Workshop on Natural Language Processing for Social Media*, Association for Computational Linguistics, Nov. 2016, pp. 48–54. doi: 10.18653/v1/W16-6208.

[6] N. Ljubešić and D. Fišer, "A Global Analysis of Emoji Usage," in *Proceedings of the 10th Web as Corpus Workshop*, Association for Computational Linguistics, Aug. 2016, pp. 82–89. doi: 10.18653/v1/W16-2610.

[7] I. Boutet, M. LeBlanc, J. A. Chamberland, and C. A. Collin, "Emojis influence emotional communication, social attributions, and information processing," *Computers in Human Behavior*, vol. 119, p. 106722, Jun. 2021, doi: 10.1016/j.chb.2021.106722.

[8] P. Gupta, A. Pandey, A. Kumar, and D. K. Vishwakarma, "Attention-free based dual-encoder mechanism for Aspect-based Multimodal Sentiment Recognition," in *2023 International Conference in Advances in Power, Signal, and Information Technology (APSIT)*, Bhubaneswar, India: IEEE, Jun. 2023, pp. 534–539. doi: 10.1109/APSIT58554.2023.10201711.

[9] A. Chhabra and D. K. Vishwakarma, "Multimodal hate speech detection via multi-scale visual kernels and knowledge distillation architecture," *Engineering Applications of Artificial Intelligence*, vol. 126, Nov. 2023, doi: https://doi.org/10.1016/j.engappai.2023.106991.

[10] A. Pandey and D. K. Vishwakarma, "Multimodal Sarcasm Detection (MSD) in Videos using Deep Learning Models," in *2023 International Conference in Advances in Power, Signal, and Information Technology (APSIT)*, Jun. 2023. doi: 10.1109/APSIT58554.2023.10201731.

[11] S. Aggarwal, A. Pandey, and D. K. Vishwakarma, "Multimodal Sarcasm Recognition by Fusing Textual, Visual and Acoustic content via Multi-Headed Attention for Video Dataset," in *2023 World Conference on Communication & Computing (WCONF)*, Jul. 2023, pp. 1–5. doi: 10.1109/WCONF58270.2023.10235179.

[12] F. Barbieri, M. Ballesteros, and H. Saggion, "Are Emojis Predictable?," in *Proceedings of the 15th Conference of the European Chapter of the Association for Computational Linguistics: Volume 2, Short Papers*, Association for Computational Linguistics, Apr. 2017, pp. 105–111. Accessed: Apr. 08, 2024. [Online]. Available: https://aclanthology.org/E17-2017

[13] Z. Ebrahimian, R. Toosi, and M. A. Akhaee, "Multinomial Emoji Prediction Using Deep Bidirectional Transformers and Topic Modeling," in *2022 30th International Conference on Electrical Engineering (ICEE)*, May 2022, pp. 272–277. doi: 10.1109/ICEE55646.2022.9827247.

[14] R. Wahyuni and I. Budi, "Combining Linguistic, Semantic and Lexicon Feature for Emoji Classification in Twitter Dataset," *Procedia Computer Science*, vol. 135, pp. 194–201, Jan. 2018, doi: 10.1016/j.procs.2018.08.166.

[15] Z. Al-Halah, A. Aitken, W. Shi, and J. Caballero, "Smile, Be Happy :) Emoji Embedding for Visual Sentiment Analysis," in *2019 IEEE/CVF International Conference on Computer Vision Workshop (ICCVW)*, Oct. 2019, pp. 4491–4500. doi: 10.1109/ICCVW.2019.00550.

[16] S. Cappallo, S. Svetlichnaya, P. Garrigues, T. Mensink, and C. G. M. Snoek, "New Modality: Emoji Challenges in Prediction, Anticipation, and Retrieval," *IEEE Transactions on Multimedia*, vol. 21, no. 2, pp. 402–415, Feb. 2019, doi: 10.1109/TMM.2018.2862363.

[17] C. Liu, T. Liu, S. Yang, and Y. Du, "Individual Emotion Recognition Approach Combined Gated Recurrent Unit With Emoticon Distribution Model," *IEEE Access*, vol. 9, pp. 163542–163553, 2021, doi: 10.1109/ACCESS.2021.3124585.

[18] G. Zhao, Z. Liu, Y. Chao, and X. Qian, "CAPER: Context-Aware Personalized Emoji Recommendation," *IEEE Transactions on Knowledge and Data Engineering*, vol. 33, no. 9, pp. 3160–3172, Sep. 2021, doi: 10.1109/TKDE.2020.2966971.

[19] S. Al-Azani and E.-S. M. El-Alfy, "Early and Late Fusion of Emojis and Text to Enhance Opinion Mining | IEEE Journals & Magazine | IEEE Xplore," *IEEE Access*, vol. 9, pp. 121031–121045, doi: 10.1109/ACCESS.2021.3108502.

[20] G. V. Singh, M. Firdaus, A. Ekbal, and P. Bhattacharyya, "Unity in Diversity: Multilabel Emoji Identification in Tweets," *IEEE Transactions on Computational Social Systems*, vol. 10, no. 3, pp. 1029–1038, Jun. 2023, doi: 10.1109/TCSS.2022.3162865.

[21] "Emoji, Sentiment and Emotion Aided Cyberbullying Detection in Hinglish | IEEE Journals & Magazine | IEEE Xplore," *IEEE Transactions on Computational Social Systems*, vol. 10, no. 5, pp. 2411–2420, Jul. 2022, doi: 10.1109/TCSS.2022.3183046.

[22] A. R. Chrismanto, A. K. Sari, and Y. Suyanto, "Enhancing Spam Comment Detection on Social Media With Emoji Feature and Post-Comment Pairs Approach Using Ensemble Methods of Machine Learning," *IEEE Access*, vol. 11, pp. 80246–80265, Jul. 2023, doi: 10.1109/ACCESS.2023.3299853.

[23] D. Peng and H. Zhao, "Seq2Emoji: A hybrid sequence generation model for short text emoji prediction," *Knowledge-Based Systems*, vol. 214, p. 106727, Feb. 2021, doi: 10.1016/j.knosys.2020.106727.

[24] S. Lee, D. Jeong, and E. Park, "MultiEmo: Multi-task framework for emoji prediction," *Knowledge-Based Systems*, vol. 242, p. 108437, Apr. 2022, doi: 10.1016/j.knosys.2022.108437.

[25] A. Gupta *et al.*, "Context-Aware Emoji Prediction Using Deep Learning," in *Artificial Intelligence and Speech Technology*, in Communications in Computer and Information Science. Springer International Publishing, 2022, pp. 244–254. doi: 10.1007/978-3-030-95711-7_22.

[26] L. Zhang, Y. Zhou, T. Erekhinskaya, and D. Moldovan, "Emoji Prediction: A Transfer Learning Approach," in *Advances in Information and Communication*, S. Kapoor, Ed., in Advances in Intelligent Systems and Computing. Springer International Publishing, 2020, pp. 864–872. doi: 10.1007/978-3-030-39442-4_65.

[27] G. Guibon, M. Ochs, and P. Bellot, "From Emoji Usage to Categorical Emoji Prediction," in *Computational Linguistics and Intelligent Text Processing*, in Lecture Notes in Computer Science. Springer Nature Switzerland, 2023, pp. 329–338. doi: 10.1007/978-3-031-23804-8_26.

[28] L. Duarte, L. Macedo, and H. Gonçalo Oliveira, "Emoji Prediction for Portuguese," in *Computational Processing of the Portuguese Language*, in Lecture Notes in Computer Science. Springer International Publishing, 2020, pp. 174–183. doi: 10.1007/978-3-030-41505-1_17.

[29] H. Asano and M. Matsuhara, "Multi-task Learning Method Using Emoji Prediction as Auxiliary Task for Sentiment Analysis," in *Proceedings of Eighth International Congress on Information and Communication Technology*, in Lecture Notes in Networks and Systems. Springer Nature, 2023, pp. 521–533. doi: 10.1007/978-981-99-3091-3_43.

[30] G. S. S. N. Himabindu, R. Rao, and D. Sethia, "A self-attention hybrid emoji prediction model for code-mixed language: (Hinglish)," *Soc. Netw. Anal. Min.*, vol. 12, no. 1, p. 137, Sep. 2022, doi: 10.1007/s13278-022-00961-1.

[31] L. Duarte, L. Macedo, and H. Gonçalo Oliveira, "Exploring Emojis for Emotion Recognition in Portuguese Text," in *Progress in Artificial Intelligence*, in Lecture Notes in Computer Science. Springer International Publishing, 2019, pp. 719–730. doi: 10.1007/978-3-030-30244-3_59.

[32] F. Barbieri, M. Ballesteros, F. Ronzano, and H. Saggion, "Multimodal Emoji Prediction," in *Proceedings of the 2018 Conference of the North American Chapter of the Association for Computational Linguistics: Human Language Technologies*, Association for Computational Linguistics, 2018, pp. 679–686. doi: 10.18653/v1/N18-2107.

[33] F. Barbieri *et al.*, "SemEval 2018 Task 2: Multilingual Emoji Prediction," in *Proceedings of The 12th International Workshop on Semantic Evaluation*, New Orleans, Louisiana: Association for Computational Linguistics, 2018, pp. 24–33. doi: 10.18653/v1/S18-1003.

[34] S. Jin and T. Pedersen, "Duluth UROP at SemEval-2018 Task 2: Multilingual Emoji Prediction with Ensemble Learning and Oversampling," in *Proceedings of The 12th International Workshop on Semantic Evaluation*, New Orleans, Louisiana: Association for Computational Linguistics, 2018, pp. 482–485. doi: 10.18653/v1/S18-1077.

[35] C. Baziotis, A. Nikolaos, A. Kolovou, G. Paraskevopoulos, N. Ellinas, and A. Potamianos, "NTUA-SLP at SemEval-2018 Task 2: Predicting Emojis using RNNs with Context-aware Attention," in *Proceedings of The 12th International Workshop on Semantic Evaluation*, New Orleans, Louisiana: Association for Computational Linguistics, 2018, pp. 438–444. doi: 10.18653/v1/S18-1069.

[36] Z. Chen, S. Shen, Z. Hu, X. Lu, Q. Mei, and X. Liu, "Emoji-Powered Representation Learning for Cross-Lingual Sentiment Classification," in *The World Wide Web Conference*, San Francisco CA USA: ACM, May 2019, pp. 251–262. doi: 10.1145/3308558.3313600.

[37] V. N. Durga Pavithra Kollipara, V. N. Hemanth Kollipara, and M. D. Prakash, "Emoji Prediction from Twitter Data using Deep Learning Approach," in *2021 Asian Conference on Innovation in Technology (ASIANCON)*, PUNE, India: IEEE, Aug. 2021, pp. 1–6. doi: 10.1109/ASIANCON51346.2021.9544680.

[38] J. Wang *et al.*, "Deep High-Resolution Representation Learning for Visual Recognition," *IEEE Transactions on Pattern Analysis and Machine Intelligence*, vol. 43, no. 10, pp. 3349–3364, Oct. 2021, doi: 10.1109/TPAMI.2020.2983686.

[39] M. De Marsico, M. Nappi, D. Riccio, and H. Wechsler, "Robust Face Recognition for Uncontrolled Pose and Illumination Changes," *IEEE Trans. Syst. Man Cybern, Syst.*, vol. 43, no. 1, pp. 149–163, Jan. 2013, doi: 10.1109/TSMCA.2012.2192427.

[40] A. Castiglione, P. Vijayakumar, M. Nappi, S. Sadiq, and M. Umer, "COVID-19: Automatic Detection of the Novel Coronavirus Disease From CT Images Using an Optimized Convolutional Neural Network," *IEEE Trans. Ind. Inf.*, vol. 17, no. 9, pp. 6480–6488, Sep. 2021, doi: 10.1109/TII.2021.3057524.

[41] M. Soleymani, D. Garcia, B. Jou, B. Schuller, S.-F. Chang, and M. Pantic, "A survey of multimodal sentiment analysis," *Image and Vision Computing*, vol. 65, pp. 3–14, Sep. 2017, doi: 10.1016/j.imavis.2017.08.003.

[42] A. Pandey and D. K. Vishwakarma, "VABDC-Net: A framework for Visual-Caption Sentiment Recognition via spatio-depth visual attention and bi-directional caption processing," *Knowledge-Based Systems*, vol. 269, pp. 1105–11015, Jun. 2023, doi: https://doi.org/10.1016/j.knosys.2023.110515.

[43] A. Pandey and D. K. Vishwakarma, "Attention-based Model for Multi-modal sentiment recognition using Text-Image Pairs," in *2023 4th International Conference on Innovative Trends in Information Technology (ICITIIT)*, IEEE, Mar. 2023. doi: 10.1109/ICITIIT57246.2023.10068626.

[44] A. Yadav and D. K. Vishwakarma, "AW-MSA: Adaptively weighted multi-scale attentional features for DeepFake detection," *Engineering Applications of Artificial Intelligence*, vol. 127, p. 107443, Jan. 2024, doi: 10.1016/j.engappai.2023.107443.

[45] A. Dosovitskiy *et al.*, "An Image is Worth 16x16 Words: Transformers for Image Recognition at Scale," presented at the ICLR 2021, ICLR, Jun. 2021.

[46] A. Vaswani *et al.*, "Attention is All you Need," in *Advances in Neural Information Processing Systems*, Curran Associates, Inc., 2017. Accessed: Mar. 31, 2024. [Online]. Available: https://proceedings.neurips.cc/paper_files/paper/2017/hash/3f5ee243547dee91fbd053c1c4a845aa-Abstract.html

[47] C. Raffel *et al.*, "Exploring the limits of transfer learning with a unified text-to-text transformer," *J. Mach. Learn. Res.*, vol. 21, no. 1, p. 140:5485-140:5551, Jan. 2020.

[48] P. Shaw, J. Uszkoreit, and A. Vaswani, "Self-Attention with Relative Position Representations," in *Proceedings of the 2018 Conference of the North American Chapter of the Association for Computational Linguistics: Human Language Technologies, Volume 2 (Short Papers)*, New Orleans, Louisiana: Association for Computational Linguistics, Jun. 2018, pp. 464–468. doi: 10.18653/v1/N18-2074.

[49] A. Radford *et al.*, "Learning Transferable Visual Models From Natural Language Supervision," in *Proceedings of the 38th International Conference on Machine Learning*, PMLR, Jul. 2021, pp. 8748–8763. Accessed: Apr. 05, 2024. [Online]. Available: https://proceedings.mlr.press/v139/radford21a.html

[50] M. Schuster and K. K. Paliwal, "Bidirectional recurrent neural networks," *IEEE Transactions on Signal Processing*, vol. 45, no. 11, pp. 2673–2681, Nov. 1997, doi: 10.1109/78.650093.

[51] A. Joulin, E. Grave, P. Bojanowski, and T. Mikolov, "Bag of Tricks for Efficient Text Classification," in *Proceedings of the 15th Conference of the European Chapter of the Association for Computational Linguistics: Volume 2, Short Papers*, Association for Computational Linguistics, Apr. 2017, pp. 427–431. Accessed: Apr. 08, 2024. [Online]. Available: https://aclanthology.org/E17-2068

[52] J. Devlin, M.-W. Chang, K. Lee, and K. Toutanova, "BERT: Pre-training of Deep Bidirectional Transformers for Language Understanding," in *Proceedings of the 2019 Conference of the North American Chapter of the Association for Computational Linguistics: Human Language Technologies, Volume 1 (Long and Short Papers)*, J. Burstein, C. Doran, and T. Solorio, Eds., Minneapolis, Minnesota: Association for Computational Linguistics, Jun. 2019, pp. 4171–4186. doi: 10.18653/v1/N19-1423.

[53] L. Zhuang, L. Wayne, S. Ya, and Z. Jun, "A Robustly Optimized BERT Pre-training Approach with Post-training," in *Proceedings of the 20th Chinese National Conference on Computational Linguistics*, Huhhot, China: Chinese Information Processing Society of China, Aug. 2021, pp. 1218–1227. Accessed: Oct. 25, 2023. [Online]. Available: https://aclanthology.org/2021.ccl-1.108

[54] K. He, X. Zhang, S. Ren, and J. Sun, "Deep Residual Learning for Image Recognition," in *2016 IEEE Conference on Computer Vision and Pattern Recognition (CVPR)*, Las Vegas, NV, USA: IEEE, Jun. 2016, pp. 770–778. doi: 10.1109/CVPR.2016.90.

[55] M. Tan and Q. V. Le, "EfficientNet: Rethinking Model Scaling for Convolutional Neural Networks." arXiv, Sep. 11, 2020. Accessed: May 22, 2024. [Online]. Available: http://arxiv.org/abs/1905.11946

[56] S. Xie, R. Girshick, P. Dollar, Z. Tu, and K. He, "Aggregated Residual Transformations for Deep Neural Networks," in *2017 IEEE Conference on Computer Vision and Pattern Recognition (CVPR)*, Honolulu, HI: IEEE, Jul. 2017, pp. 5987–5995. doi: 10.1109/CVPR.2017.634.

[57] I. Radosavovic, R. P. Kosaraju, R. Girshick, K. He, and P. Dollar, "Designing Network Design Spaces," in *2020 IEEE/CVF Conference on Computer Vision and Pattern Recognition (CVPR)*, Seattle, WA, USA: IEEE, Jun. 2020, pp. 10425–10433. doi: 10.1109/CVPR42600.2020.01044.